# A SUCCESSFUL INTEGRATION OF THE ROBOTIC TECHNOLOGY KERNEL (RTK) FOR A BY-WIRE ELECTRIC VEHICLE SYSTEM WITH A MOBILE APP INTERFACE

**Justin Dombecki[1], James Golding[1], Mitchell Pleune[1], Nicholas Paul[1], Chan-Jin Chung[1]**

[1]Computer Science Autonomous Robotics (CAR) lab, Lawrence Technological University, Southfield, MI

## ABSTRACT

*We were able to complete the full integration of the Robotic Technology Kernel (RTK) into an electric vehicle by-wire system using lidar and GPS sensors. The solution included a mobile application to interface with the RTK-enabled autonomous vehicle. Altogether the system was designed to be modular, using the concepts of message-based software design that is built into the Robot Operating System (ROS), which is at the foundation of RTK. The team worked incrementally to develop working software to demonstrate each milestone on the path to successfully completing the RTK integration for the development of an application called the Vehicle Summoning System (VSS).*

## 1. INTRODUCTION

As one of the Autonomous Campus Transport (ACTor) research projects [1], the team exercised a six-iteration approach to achieve their goals, of a second full stack autonomous vehicle – this time, built on top of the evolved Robotic Technology Kernel (RTK) [2, 3] with an additional mobile phone application interface. Learned best practices from the original ACTor1 project were employed to assemble and configure the necessary sensors and computer hardware on the second electric vehicle (ACTor2, see Figure 0). Starting with the setup of a System Integration Lab (SIL) [4], the software solution was designed to be modular and reusable, which are concepts at the core of the messaged-based Robotic Operating System (ROS), and its existing integrations with RTK [5].

The vehicle, a Polaris Gem 2, was configured with a drive by-wire system (DBW) [6] from Dataspeed Inc., providing a level of abstraction above the mechanization components of the vehicle. Interfaces, using ROS nodes, were provided to the research team with which to integrate. These, and other factors, closely resembled the ACTor1 vehicle enabling the team to work from the group experience and begin writing software early using a vehicle simulator, while the ACTor2 by-wire system was still being configured.

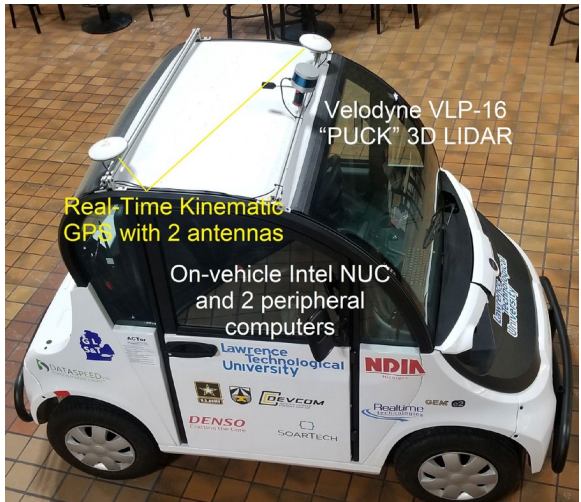

Figure 0: ACTor2 vehicle

The end-product was a system of both hardware and software. The hardware included an on-vehicle Intel NUC and two peripheral computers (a laptop and a mobile phone), which connect to the on-vehicle computer through a Wi-Fi connection, to execute commands from outside of the vehicle. The software solution included four ROS nodes (discussed in section 4) and a mobile application for the Android OS.

This paper introduces the six development iterations in detail in section 2. Section 3 explains how we finalize the vehicle hardware and sensor configuration for the RTK integration. In Section 4, we describe designed resultant system architecture. Demonstration of developed functions are explained in Section 5. Summary of the results is described in Section 6.

## 2. INCREMENTAL DEVELOPMENT METHODOLOGY

Given the applied research on the first autonomous vehicle (ACTor1), the team was able to incrementally configure the hardware and software components needed for the RTK integration on ACTor2.

Additionally, this research was affected by an unprecedented lockdown due to the global pandemic in early 2020. To continue, while following the guidelines from the Centers for Disease Control and Prevention (CDC), the team chose to meet online-only and changed their planned iteration goals to accommodate the limitation. As the research continued, the restrictions evolved, and the team was able to work together in-person again.

The main goals of the six iterations:
(1) Enable vector localization (position and heading) of the vehicle, by implementing a Real-Time Kinematic Global Positioning System (RTK-GPS) with two antennas on ACTor1.
(2) Enable the System Integration Lab (SIL), by integrating the Lawrence Tech campus map on the Warfighter Machine Interface (WMI). Tested end-to-end WMI features with the ANVEL vehicle simulator.
(3) Integrate a Proof-of-Concept (POC) of the Vehicle Summoning System mobile application (VSS) and the ROS node used to bridge between the phone and vehicle, *ltu-taxi*. The ANVEL vehicle simulator would be used for testing.
(4) Mechanize the new autonomous vehicle, ACTor2, with sensors and computers.
(5) Ensure the functionality of end-to-end VSS – by testing with an Android phone and the ACTor2 vehicle running the *ltu-taxi* node.
(6) Complete the integration of RTK with the ACTor2 sensors and the drive-by-wire system. The WMI would be used to validate functionality and system status.

These iterations were realized organically through the Agile process of understanding limitations and risks, against which to build feasible goals of demonstrable working software [7]. The team was able to incorporate the 'responding to change' Agile principle, among others, by evolving their approach – to complete the project scope through quarantine and strict in-person social distancing and mask mandates.

When the research team found issues or opportunities for improvement, the finding was documented and reviewed to determine its priority within the project scope. These interactions between individuals yielded a second version of the mobile application, vast improvements to the accuracy of route following and the procurement of additional networking components to ensure system performance.

## 3. THE RTK-INTEGRATED VEHICLE

Building on the research from the ACTor project, the team was able to both use the vehicle in this research and reference its hardware and software configuration to streamline the build out of ACTor2. As mentioned, before the second vehicle was configured and ready for integration, the team began researching a new concept on ACTor1 called vector localization. This required two GPS antennas to determine the heading and position of the vehicle, using a concept called Real-Time Kinematic GPS (RTK-GPS) [8, 9]. Previously, ACTor1 had Single Point Positioning (SPP) using a single GPS antenna.

Understanding the focused scope required for a RTK integration, the team chose to mount only a lidar and an RTK-GPS system to ACTor2. The lidar was a Velodyne VLP-16 "PUCK" 3D lidar. The RTK-GPS system consisted of a Piksi Multi Global Navigation Satellite System (GNSS) using the Compass configuration and a stationary base, available from Swift Navigation. With this set of sensors, the team was able to integrate with the *world_model* and *localization* ROS nodes from RTK, eventually integrating with the path planner engines, Maverick and Vaquerito.

Drawing on lessons learned from ACTor1, the team configured a similar computer and networking approach, using a network access point connected to a switch along with all other peripherals to a central processing computer, which would send ROS *Twist* messages to the by-wire system. With each software component assigned an IP and physical measurements made, the team assembled the *actor2_rtk_launcher* ROS node, which holds the ACTor2 configuration dataset and launches RTK with the appropriate parameters.

## 4. SYSTEM ARCHITECTURE DESIGN

At the highest-level context, the system design architecture is defined by five ROS subsystems: native Lidar nodes, native GPS nodes, native drive-by-wire nodes, RTK nodes and the nodes created by the team at LTU. These five subsystems are shown in figure 1.

To further discuss the integration points of these subsystems, four functionalities can be extracted and outlined.

### *4.1 Integration of RTK Driving*

The integration of RTK driving is made possible by the *speed_curv_to_twist* custom ROS node. This node converts the *vehicle_interface* RTK messages from a speed and curvature context to the ROS *Twist* message, which is the message type necessary to interface with the drive-by-wire system (DBW). The ROS topic available from the *dataspeed_ulc_can* DBW node is the *polaris/vehicle/cmd_vel* ROS topic. Figure 2 shows the path of messages from the ROS topics, *steering_sense*, *speed_setpoint*, and *curvature_setpoint*, within the RTK context, *vehicle_interface*, to the LTU ROS custom node *speed_curv_to_twist* from the *actor2_rtk_support* package.

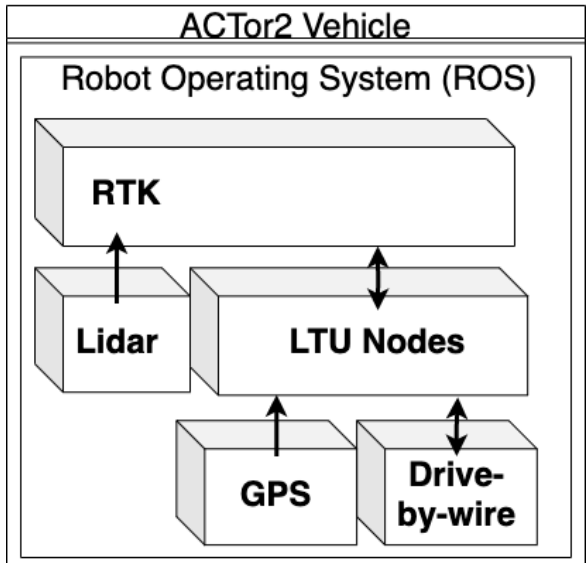

Figure 1: High Level Vehicle Software Architecture

### 4.2 Integration of Sensors

Each sensor takes its own path to integrate with RTK, as shown in figure 3. The lidar subsystem integrates directly with RTK, as the *velodyne* ROS node is supported by the RTK *world_model* context. The lidar points are communicated over the *velodyne_points* topic, which is configured in the RTK configuration node, *actor2_rtk_launcher*.

The GPS, however, requires middleware integration to take advantage of the precision afforded by the Real-Time Kinematic GPS (RTK-GPS) vector localization. The LTU ROS node *piksi_odom_pub* was developed against ACTor1 during the first iteration and made available in the *actor2_support* ROS package. This node subscribes to topics from the Swift Navigation node *ethz_piksi_ros* to generate a Odometry message defined by the *nav_msgs* library. The three topics from Piksi are *enu_pose_fix* (RTK-GPS fix), *enu_pose_spp* (SPP fix) and *baseline_heading* (heading). The *piksi_odom_pub* node, then, publishes the odometry message to the */odom* topic. To interface with RTK, the *odom_repub* node of the *actor2_rtk_support* package was needed to republish the odometry message to the *localization* RTK context -- that is, the topics */near_field_odom* and */far_field_odom* in that RTK context. In addition, the LTU ROS node *actor_localization* was developed, in a later iteration, to further interpret the piksi GPS data and determine the best GPS fix for the vehicle, including SPP fallback. This node subscribes to the same */odom* topic and subscribes to the *navsatfix_best_fix* topic from *ethz_piksi_ros*.

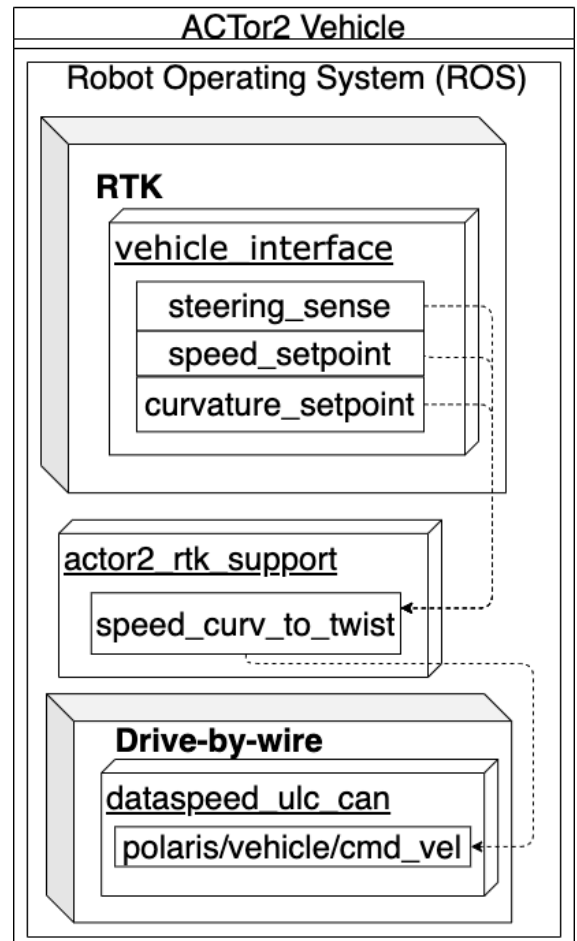

Figure 2: Path from RTK to DBW

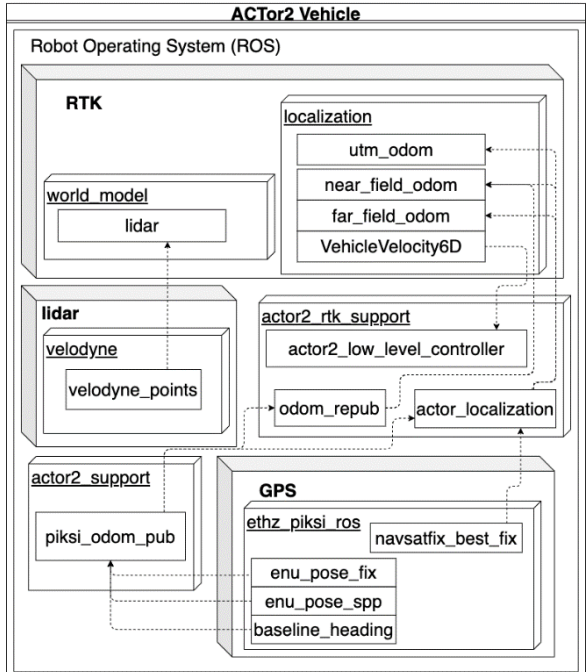

Figure 3: Paths of Sensors

It publishes to the same RTK *localization* context topics as *odom_repub* with the addition of */utm_odom*. The resulting velocity matrix from the RTK contexts, *world_model* and *localization*, is the *VelocityVelocity6D* topic, to which the *actor2_low_level_controller* subscribes.

## 4.3 Integration of Vehicle Status and Estop

The integration of the overall status of the vehicle ensures a complete connection between the RTK and DBW subsystems. Additionally, safety measures were put into place to ensure the appropriate override behavior when engaging the physical e-stop or when the safety driver manually engages the steering wheel or petals. This functionality is made available by two custom ROS nodes in the *actor2_rtk_support* LTU ROS package and their interactions are detailed in figure 4. The *estop_heatbeat* node subscribes to the DBW topic *polaris/vehicle/ulc_report* to generate messages for the RTK *can* safety monitor topics, *safety_monitor_status* and *safety_monitor_estop*, and the estop-specific topic within RTK: *estop_sense* in the context of *vehicle_interface*. The lowest-level LTU ROS node, the *actor2_low_level_controller*, subscribes to this *estop_sense* RTK topic along with other *vehicle_interface* topics to maintain the vehicle state and, ultimately, the connection between the RTK and DBW subsystems.

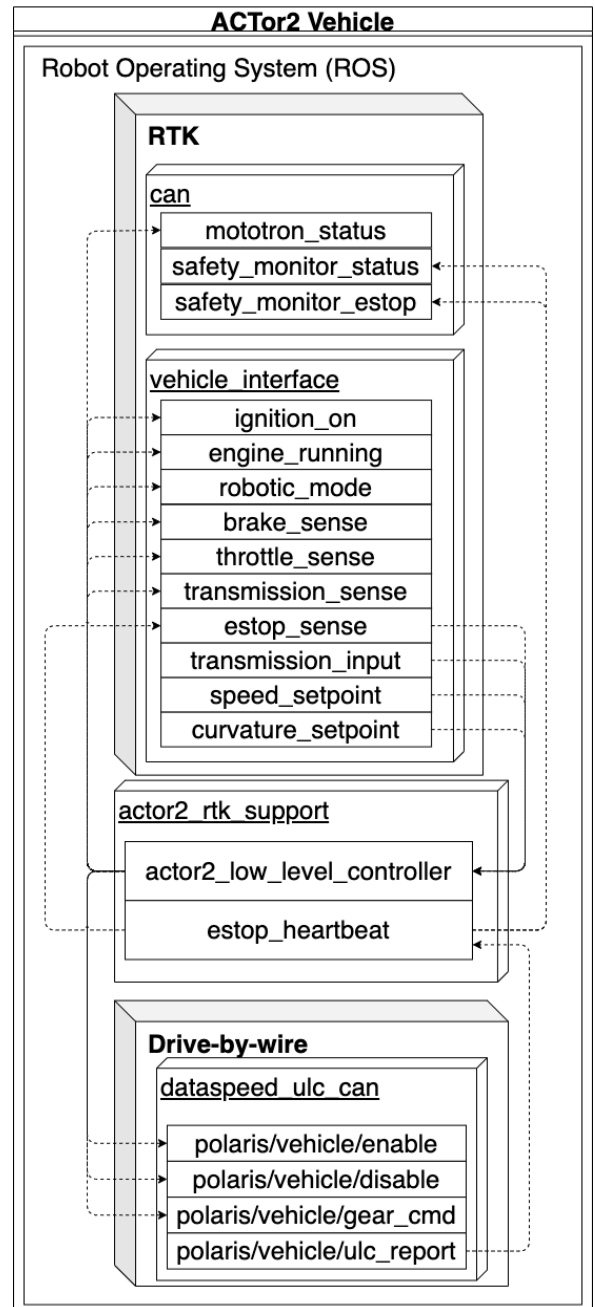

Figure 4: Paths of Vehicle Status and Estop Override

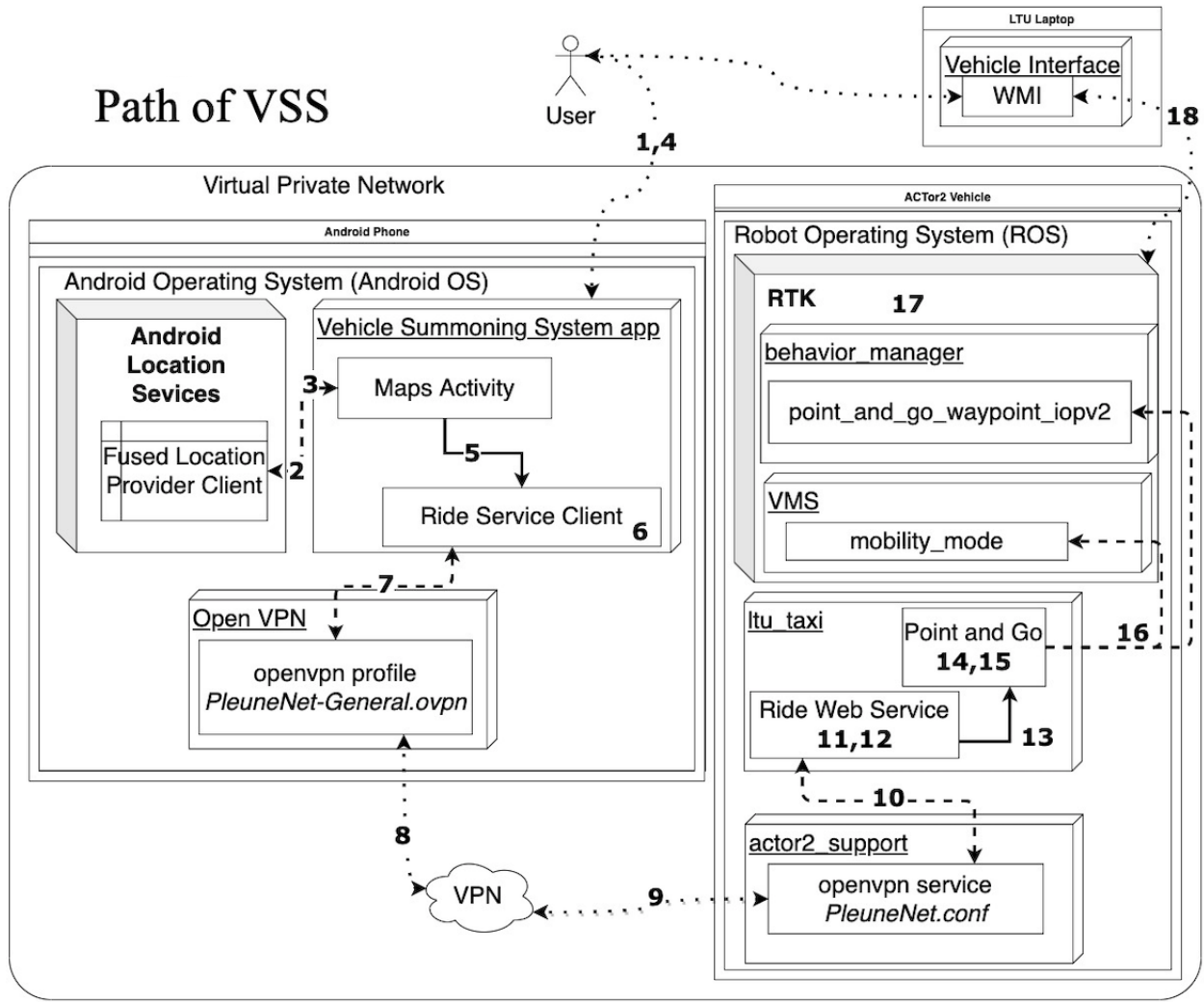

Figure 5: Path of Vehicle Summon System (VSS)

### 4.4 Integration of VSS

The integration of the Vehicle Summoning System (VSS) was first performed against the ANVEL vehicle simulator and later integrated into the ACTor2 RTK system. In both cases, the course of a signal through the system is the same. Following the eighteen steps shown in Figure 5, from a user interaction with a mobile phone to the setting of the RTK destination waypoint, one can understand the benefits of reuse from the message-based design architecture.

1) The mobile phone user launches the Vehicle Summoning System (VSS) application.

2) Using Android Location Services, the *MapsActivity* verifies that the location permissions have been granted with the *FusedLocationProviderClient*.

3) Once permissions are granted, the *MapsActivity* initiates a request to the client to receive location updates at the default frequency.

4) The user presses the 'Summon' button on the upper left corner on the VSS phone application screen.

5) The *MapsActivity* creates an instance of the *RideServiceClient* to request the

posting of the last known location to the *RideWebService*, hosted on the ACTor2.

6) The *RideServiceClient* creates a HTTP POST to the static IP address of the *RideWebService* using the standard *HttpUrlConnection* found in the Java networking framework, bundling the last known device location within a JSON string with latitude and longitude key-value pairs.

7) The HTTP POST request is sent by the *RideServiceClient* within the Java networking framework, interacting with the OpenVPN application running on the Android device.

8) The OpenVPN application routes the HTTPS POST request through the *PleuneNet-General.ovpn* configuration, subsequently, sending the request through the private-VPN hosted on Amazon Web Services (AWS).

9) The request makes its way to the host of the VPN on Actor2, a Linux service running the OpenVPN software with the *PleuneNet.conf* configuration.

10) OpenVPN routes the request to the web server, which is running within the ROS node *RideWebService*, at the address 192.168.99.5:8642.

11) The *RideWebService* node parses the JSON request containing the last known location of the Android device and returns a 200 'success' response over the HTTPS connection.

12) The location is repackaged as a custom ROS type called *PointAndGo*, containing the latitude, longitude and a mobility mode set to "go to waypoint".

13) The *RideWebService* node publishes the *PointAndGo* message on the */ltu/point_and_go/* ROS topic. The subscribing ROS node *PointAndGo* receives the message.

14) The *PointAndGo* node converts the latitude and longitude to the Universal Transverse Mercator (UTM) coordinate system creating a simple Waypoint SUMET (Small Unit Mobility Enhancement Technology) navigation message.

15) The *PointAndGo* node parses the mobility mode from the message. When "go to waypoint" is matched, a Mobility Mode dsat common message is instantiated with the 'mobilityMode' value set to '14' and the 'idleReason' value set to '0'.

16) The *PointAndGo* node publishes the Mobility Mode message to the */vms/command_mobilitymode* RTK ROS topic and the Waypoint to the */behavior_manager/point_and_go_waypoint_iopv2* RTK ROS topic.

17) RTK subscribes to these topics within the *vms* and *behavior_manager* contexts, respectfully. Using the Maverick path planner, RTK creates a route to the waypoint and updates the *vehicle_interface*.

18) The Warfighter Machine Interface (WMI) receives updates of the vehicle location and objects detected by the LIDAR, showing the movement of the vehicle on the LTU map, as it autonomously navigates to the device location.

From the interactions discussed in the integration of RTK driving, one can continue to follow the updates to the *vehicle_interface* RTK context and further, still, through the

*speed_curv_to_twist* LTU ROS node, until it reaches the ACTor2 DBW subsystem.

Later, after further testing, a Wi-Fi hotspot was added to the system to improve the reliability of the internet connection between the ACTor2 vehicle and the mobile device, improving the overall performance of the Vehicle Summoning System (VSS).

## 5. WORKING SOFTWARE DEMONSTRATIONS

Throughout the course of this research, the team has generated a catalog of videos demonstrating iterations of the system as it has been developed, documenting portions of the journey of successfully integrating RTK into an electric vehicle by-wire system. Containing much of the quarantine iteration during the development of the Vehicle Summoning System (VSS) while working off-campus, these working software demonstrations show the progression of the solution in pieces. Later, working software demonstrations using ACTor2 (8, 9 and 10) show the fully integrated RTK with an electric vehicle by-wire system.

1] Dombecki, J. (4/22/2020). HTTP Server Test. Terminal curl command makes a HTTPS POST request containing GPS coordinates. The *RideWebService* HTTP server receives the coordinates and prints them to the console and a file named *posted.json*. Steps 11 and 12 in Figure 5.
https://youtu.be/2KdxFn7U-3E

2] Golding, J. (5/6/2020). Send Vehicle to Location Test. Running the ANVEL RTK vehicle simulator and the custom *ltu_taxi* ROS package, which publishes ROS topics containing the GPS coordinates and a mobility mode command. RTK receives the published topics and moves the vehicle to request location relative to the simulator RTK vehicle in ANVEL. Steps 11 – 18 in Figure 5 (substituting the WMI with the relative movement of the RTK simulated vehicle within the ANVEL environment.)
https://youtu.be/8p-4NYZoOI4

3] Dombecki, J. (5/8/2020). Phone App Integration Test. The GPS location of the Android emulator is set to the LTU campus. The user launches the VSS Android application, grants location permission and presses the 'come to me' button, displayed by the *MapsActivity*. Still within the VSS app, the GPS coordinates are collected from the *FusedLocationProviderClient*. Then, the app makes a HTTP POST request with the GPS coordinates. The HTTP server, which is running the *RideWebService* on the same computer, receives the JSON coordinates printing them to the console. Steps 1 – 12 in Figure 5 skipping VPN integration (steps 8 – 10).
https://youtu.be/eKLnex0fqVw

4] Dombecki, J.; Golding, J. (5/13/2020). Phone-Vehicle over remote VPN Test. Revisiting the Phone App Integration test (3) and adding VPN and a second remote computer. Again, the user interacts with the VSS Android app to collect the GPS coordinates and make an HTTP POST request sending the JSON coordinates to the *RideWebService*, this time, running on a second remote machine. The request travels through a cloud VPN server using the OpenVPN Android application on the emulator and OpenVPN configuration software running on the second remote machine. Steps 1 – 12 in Figure 5.
https://youtu.be/WFoVNufqWlY

5] Dombecki, J. (5/13/2020). Vehicle Integration Test. Revisiting the Phone-Vehicle over remote VPN test (4) and including ANVEL integration. Running the *ltu_taxi* ROS node, the ANVEL RTK vehicle simulator and the VSS Android app in an emulator, the user interacts with the VSS Android app to collect the GPS coordinates and make an HTTP POST request sending the JSON coordinates to the *RideWebService* ROS node, which translates the coordinates from the JSON to the ROS *PointAndGo.msg* format. Then, the *RideWebService* ROS node publishes the coordinates to *point_and_go* ROS node, which translates the coordinates to ROS topics containing the GPS coordinates and a mobility mode command. The simulated RTK vehicle receives the messages and moves to the requested location. Steps 1 – 18 in Figure 5 (substituting the WMI with the relative movement of the RTK simulated vehicle within the ANVEL). https://youtu.be/uTtpAL6yiA0

6] Golding, J. (May 2021). ACTor2 demonstrating the Vaquerito navigation of the Robotic Technology Kernel via the Route Following function of the Warfighter Machine Interface. https://youtu.be/15SzvzB8yYY

7] Golding, J. (May 2021). ACTor2 demonstrating the Maverick navigation of the Robotic Technology Kernel via the Point-and-Go function of the Warfighter Machine Interface. https://youtu.be/fyt6htYBe2o

8] Dombecki, J., Golding, J. (May 2021). ACTor2 demonstrating the Maverick navigation of the Robotic Technology Kernel via the Vehicle Summoning System from Lawrence Technological University. https://youtu.be/PWt3tnMqb8M

## 6. RESULTS AND SUMMARY

The result of the RTK integration was successful. The research group responded to change and created new plans to overcome the challenges made by quarantine, social distancing and mask mandates. No matter the barrier, the team was persistent and driven by results. Throughout the project, they met weekly (physically or remotely) to review any updated requirements and demonstrate any working software produced. As such, many demonstration videos are available showing the progression of the Vehicle Summoning System solution (VSS) and the integration of RTK features.

For VSS, the product of the quarantine iteration was created with a modular design using the message-based concept. First, it was developed against the ANVEL vehicle simulator. Later, it was integrated into ACTor2 without any code changes. It, therefore, was end-to-end tested early during the RTK integration and continued to be a useful testing data point as the ACTor2-RTK configuration moved toward its final state.

## 7. REFERENCES

[1] N. Paul, M. Pleune, CJ. Chung, B. Warrick, S. Bleicher, C. Faulknerk (2018). "ACTor: A Practical, Modular, and Adaptable Autonomous Vehicle Research Platform". In *2018 IEEE International Conference on Electro/Information Technology (EIT)*, 10.1109/EIT.2018.8500202.

[2] J. Cymerman, S. Yim, D. Larkin, K. Pegues, W. Gengler, S. Norman, Z. Maxwell, N. Gasparri, M. Pollin, C. Calderon, J. Angle, J. Collier, "Evolving the Robotic Technology Kernal to Expand Future Force Autonomous Ground Vehicle Capabilities," In *Proceedings of the Ground Vehicle Systems Engineering and Technology Symposium* (GVSETS), NDIA, Novi, MI, Aug. 13-15, 2019.

[3] D. Pirozzo, J.P. Hecker, A. Dickinson, T. Schulteis, J. Ratowski, and B. Theisen, "Integration of the Autonomous Mobility Appliqué System into the
A Successful Integration of the Robotic Technology Kernel (RTK) for a By-Wire Electric Vehicle System with a Mobile App Interface

Page 9 of 10